\documentclass[letterpaper]{article} 
\usepackage{aaai2026}  
\usepackage{times}  
\usepackage{helvet}  
\usepackage{courier}  
\usepackage[hyphens]{url}  
\usepackage{graphicx} 
\usepackage[numbers,square,comma,sort&compress]{natbib}
\usepackage{caption} 
\usepackage{algorithm}
\usepackage{algorithmic}
\usepackage{multirow}
\usepackage{booktabs}
\usepackage{subcaption}
\usepackage[table]{xcolor}
\definecolor{LightGray}{gray}{0.95}
\definecolor{LightBlue}{rgb}{0.90,0.95,1.0}
\usepackage{newfloat}
\usepackage{listings}
\DeclareCaptionStyle{ruled}{labelfont=normalfont,labelsep=colon,strut=off} 
\floatstyle{ruled}
\newfloat{listing}{tb}{lst}{}
\floatname{listing}{Listing}
\title{Neighbor-Aware Token Reduction via Hilbert Curve for Vision Transformers}
\author{
    Yunge Li, Lanyu Xu
}
\affiliations{
Department of Computer Science and Engineering\\
Oakland University \\
Rochester, MI 48309, USA\\
$\{yungeli, lxu\}@oakland.edu$
}

\usepackage{bibentry}
\usepackage{amsmath}

\begin{document}

\maketitle

\begingroup
\renewcommand{\thefootnote}{}
\footnotetext{Code available at: \url{https://github.com/Yunge6666/NAP-MAT}}
\endgroup

\begin{abstract}
Vision Transformers (ViTs) have achieved remarkable success in visual recognition tasks, but redundant token representations limit their computational efficiency. Existing token merging and pruning strategies often overlook spatial continuity and neighbor relationships, resulting in the loss of local context. This paper proposes novel neighbor-aware token reduction methods based on Hilbert curve reordering, which explicitly preserves the neighbor structure in a 2D space using 1D sequential representations. Our method introduces two key strategies: Neighbor-Aware Pruning (NAP) for selective token retention and Merging by Adjacent Token similarity (MAT) for local token aggregation. Experiments demonstrate that our approach achieves state-of-the-art accuracy-efficiency trade-offs compared to existing methods. This work highlights the importance of spatial continuity and neighbor structure, offering new insights for the architectural optimization of ViTs.
\end{abstract}

\section{Introduction}\label{sec:intro}
In recent years, Vision Transformer (ViT)~\cite{dosovitskiy2020image} has become a mainstream method in the field of computer vision and has achieved excellent performance in multiple tasks such as image classification, object detection, and image segmentation. This is mainly due to its built-in global attention mechanism. However, global attention also brings huge computational overhead. As the input image resolution increases or the number of tokens increases, the computational complexity will increase quadratically with the number of tokens, which has become a significant problem in the practical application of ViT~\cite{khan2022transformers}.
One practical approach to reducing the computational burden is to decrease the number of tokens involved in computation. Currently, the primary token reduction methods include token pruning and token merging. The combination of these two approaches has also been attempted to reduce the token number and computational overhead further while preserving model accuracy.

In token pruning, importance scores of tokens are typically computed to determine which tokens should be retained and which should be discarded. Discarded tokens either participate minimally or do not participate in subsequent computations. Various methods have been proposed to compute token importance, such as feature value~\cite{tang2022patch}, attention scores~\cite{liang2022not,fayyaz2022adaptive,xu2022evo}, gradient-based metrics~\cite{rao2021dynamicvit,li2022sait}, energy~\cite{wang2022vtc}, or combinations thereof~\cite{ye2025fit}. Most of these methods focus exclusively on individual token characteristics, overlooking the neighboring context of tokens. To capture this critical neighboring information, this work proposes a neighbor-aware pruning method, incorporating the influence of neighboring tokens into importance evaluation to achieve comprehensive token importance scoring.

In contrast to token pruning, token merging reduces the number of tokens by merging multiple similar tokens into a single representative token. This process involves computing similarity between tokens using various metrics, including token feature vectors or attention-related features such as queries and keys. Tokens with high similarity are merged and continue to participate in subsequent computations. Nevertheless, existing studies~\cite{chen2023diffrate,lee2024learning,yuan2024efficient,choi2024vid} typically calculate token similarity on a global scale, ignoring an essential characteristic of image data: pixel changes in images tend to be gradual and continuous~\cite{burt1987laplacian}. Therefore, this work proposes a localized token merging strategy that computes similarity and performs merging operations exclusively within local, adjacent token regions.
\begin{figure}[htp]
    \centering
    \includegraphics[width=1.0\linewidth]{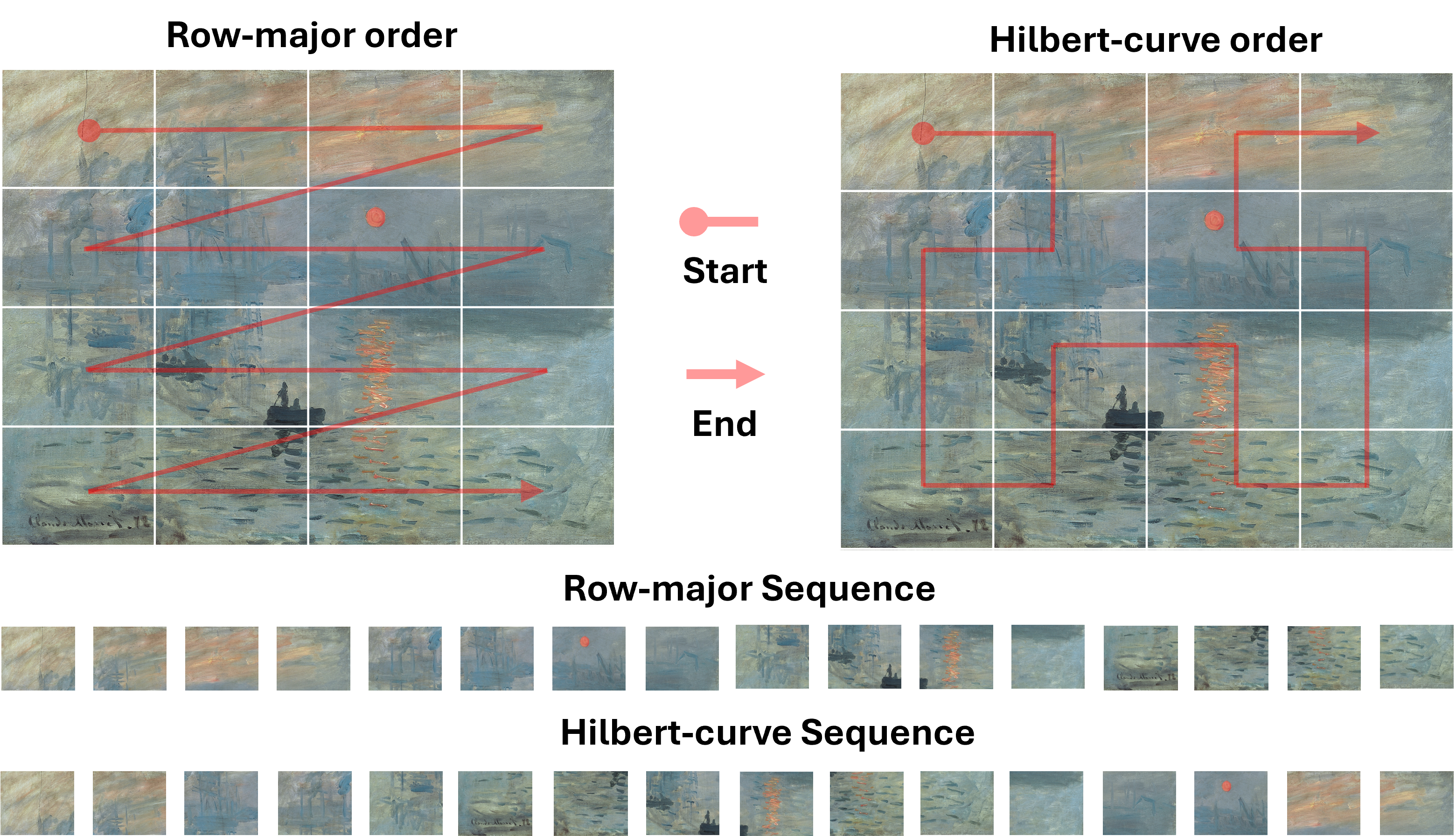}
    \caption{Token order. Illustration of token sequence reordering from row-major order (left) to Hilbert curve order (right), which better preserves spatial locality in the 1D token sequence. The reordered tokens are flattened below, demonstrating how Hilbert-curve ordering brings adjacent image patches closer in the 1D sequence.}
    \label{fig:sequence}
\end{figure}

The proposed methods in the work, whether pruning or merging tokens, consistently highlight the significance of local spatial context in token reduction. Although some studies have proposed a variety of local partitioning strategies, such as standard window partitioning in Swin Transformer~\cite{liu2021swin} and horizontal and vertical axis region partitioning in CSwin~\cite{dong2022cswin}, these methods cannot be directly and effectively applied to token reduction. In addition, some clustering-based region partitioning algorithms~\cite{mittal2022comprehensive} may also be used for local partitioning, but these methods usually significantly increase the additional computational burden. The fundamental challenge lies in preserving 2D neighboring structures when flattening images into 1D token sequences. Since the Hilbert curve provides an optimal mapping from 2D to 1D while maintaining spatial locality, this work proposes to use the Hilbert curve to reorder the image patches. As shown in Figure~\ref{fig:sequence}, compared with the row-major order in the original ViT, the reorder of the Hilbert curve ensures that adjacent tokens in the 2D space remain adjacent in the 1D sequence. It should be noted that the Hilbert curve reorder only changes the order of the tokens in the 1D sequence, and does not have any substantial impact on the subsequent attention calculation, so it is easy to implement. After the Hilbert curve reorder, the local neighboring areas in the token reduction can be more easily identified and processed, and all calculations are performed on the reordered 1D sequence without additional consideration of the 2D spatial position of the token.
In summary, the main contributions of this work include:
\begin{itemize}
    \item \textbf{Novel application of Hilbert curves}: We innovate the use of space-filling Hilbert curves for token reduction in ViT and its variants, effectively maintaining 2D spatial relationships within a 1D sequence.
    \item \textbf{Merging by Adjacent Tokens (MAT)}: We propose an efficient local merging strategy that aggregates tokens based on their spatial adjacency and feature similarity.
    \item \textbf{Neighbor-Aware Pruning (NAP)}: We introduce a pruning method that preserves context by considering both token attention and neighborhood relationships.
\end{itemize}


\section{Related Work}\label{sec:related}
\subsection{Hilbert Curve}
David Hilbert proposed the Hilbert curve in 1891~\cite{hilbert1935stetige}. It defines a mapping relationship between one-dimensional and two-dimensional space, which can effectively preserve the spatial locality of data. In recent years, the Hilbert curve has been extended to 3D data processing tasks such as disordered point clouds. For example, the Hilbert curve is used in Point Transformer v3~\cite{wu2024point} to replace the traditional K-nearest neighbor (KNN) method, thereby more efficiently defining the local neighborhood and facilitating local feature extraction. In addition, Kutscher et al. explored the application of Hilbert curves and other space-filling curves in various vision models~\cite{kutscher2025reordering}. Their experimental results show that reordering the token sequence of ViT according to the Hilbert curve has a limited effect on the performance of the model. Our analysis suggests that Hilbert curve reorder provides better spatial proximity for the token sequence, which will be beneficial for token reduction tasks.
\subsection{Token Pruning}
The basic idea of token pruning is to reduce the number of tokens involved in subsequent calculations by evaluating and discarding unimportant tokens, thereby reducing the computational overhead of the model. DynamicViT~\cite{rao2021dynamicvit} is a classic token pruning method that uses a lightweight dynamic prediction module to decide whether to keep or remove tokens in each layer, significantly improving computational efficiency. EViT~\cite{chen2023sparsevit} directly uses the attention scores from the classification token (CLS token) to other tokens to evaluate their importance. Important tokens are retained, while unimportant tokens are discarded or merged to participate in subsequent calculations. A-ViT~\cite{yin2022vit} calculates a halting probability for each token to determine whether it should continue to be passed to deeper layers of the network. A-ViT cleverly leverages the existing MLP module in the Transformer to compute the halting probability, thereby avoiding additional parameters and computation. In addition, recent methods such as SparseViT~\cite{chen2023sparsevit} and SPViT~\cite{kong2022spvit} have also explored token pruning, further improving pruning efficiency and effectiveness.
\subsection{Token Merging}
The main idea of token merging is to reduce the number of tokens and computational overhead by merging multiple similar tokens into a single token. Bolya et al. proposed the Token Merging (ToMe)~\cite{bolya2022token}. This method first calculates the cosine similarity between tokens and uses the Bipartite Soft Matching (BSM) strategy to select the merged token pairs, and finally merges them in a weighted average manner. Subsequently, this method was successfully extended and applied to Stable Diffusion to accelerate image generation~\cite{bolya2023token}.

In addition, some studies have attempted to combine token pruning and token merging strategies to further improve the effectiveness of token reduction. DiffRate uses a pruning method similar to EViT to prune tokens according to their importance, and then calculates the similarity of the retained tokens and merges them, thereby further reducing the number of tokens. The ratios of pruning and merging in DiffRate are both learnable parameters~\cite{chen2023diffrate}. Wei et al. proposed Token Pruning $\&$ Squeezing (TPS), which combines attention-based token pruning with lightweight MLP-based token merging, adaptively adjusting compression rates per layer via a learnable gate~\cite{wei2023joint}.

\begin{figure*}[htp]
    \centering
    \includegraphics[width=0.97\linewidth]{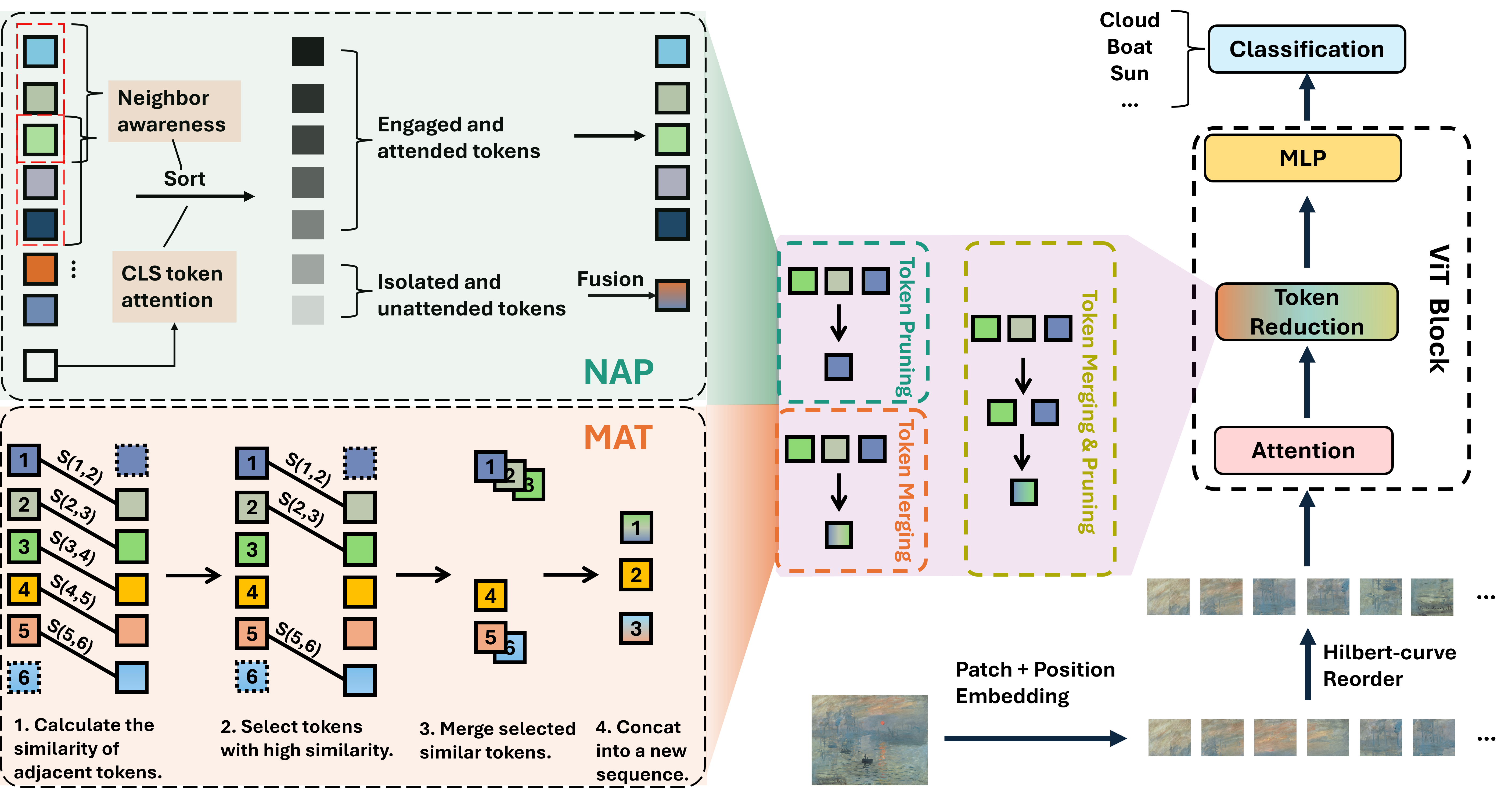}
    \caption{\textbf{Pipeline of Hilbert Curve–based Neighbor‑Aware Token Reduction.} After the image is patched, it is reordered by the Hilbert curve and then sent to the ViT block. In each ViT block, a token reduction module is inserted between the attention and MLP modules to reduce the number of tokens. The specific reduction method can be token pruning (NAP), token merging (NAP), or a combination of the two. After the token reduction module, the computational burden of subsequent modules and blocks is reduced.}
    \label{fig:overview}
\end{figure*}

In general, existing token reduction methods primarily rely on individual token attributes, such as feature representations or attention weights, to determine importance or similarity, while often overlooking the spatial context of neighboring tokens. In this work, we explicitly introduce a neighbor-awareness mechanism into the token reduction. This allows the token reduction to exploit local spatial relationships among tokens better, leading to more context-aware token selection.

\section{Neighbor-Aware Token Reduction}\label{sec:tokenredcution}
The current mainstream approaches to token reduction primarily include token pruning, token merging, and their combination. In this work, instead of relying solely on the attributes of individual tokens, we also consider the surrounding context, specifically the information provided by neighboring tokens. This neighbor-aware strategy is independently applied to both token pruning and token merging.

The overall pipeline is illustrated in Figure~\ref{fig:overview}. Specifically, the input image is first divided into a set of patches, which are then reordered according to the Hilbert curve. This reordering enables the resulting 1D token sequence better to preserve the spatial locality of the original 2D image. It also provides the structural basis for introducing neighbor-aware token reduction in later stages. Notably, the attention computation remains unaffected, since the attention mechanism in ViTs is inherently global. Based on the resulting attention weights, token reduction is applied to selectively reduce the number of tokens, either by pruning less informative tokens or merging similar ones. After passing through the token reduction module, the total number of tokens is decreased, thereby reducing the computational cost of the subsequent modules and blocks.

\subsection{Token Pruning: NAP}
Token pruning aims to remove unimportant tokens from subsequent computation entirely, or alternatively, to aggregate all unimportant tokens into a single representative token. In this work, we introduce NAP (Neighbor‑Aware Token Pruning). The process is illustrated as the NAP module in Figure~\ref{fig:overview}. The main idea is to evaluate token importance by combining each token’s attention weight from the CLS token and the influence of its neighbors. Based on the evaluated importance score, unimportant tokens are either removed entirely or merged into a single representative token.

Specifically, NAP calculates the importance score as follows. After the attention computation is completed, NAP first extracts the attention received by each token to calculate the \textit{average received attention} across all $token \rightarrow token$ pairs, as shown in Equation~\ref{rattention}. For each sample in the batch indexed by $b$, and for each non-class token indexed by $i$ (where $i = 1, \dots, M$ and $M = N - 1$, with $N$ being the total number of tokens including the class token), the attention score ${\mathrm{R\_attn}_{b,i}}$ is computed by first averaging the attention weights over all query tokens $q = 0, \dots, N-1$ for each attention head $h = 1, \dots, H$, and then averaging across all $H$ heads. Here, $\mathrm{attn}_{b,h,q,i+1}$ denotes the attention weight from query token $q$ to token $i+1$, where the index is shifted by one to skip the class token at position 0.
\begin{align}
    \
    \mathrm{R\_attn}_{b,i}
    = \frac{1}{H}
      \sum_{h=1}^{H}
        \Biggl(
          \frac{1}{N}
          \sum_{q=0}^{N-1}
            \mathrm{attn}_{b,h,q,i+1}
        \Biggr)
    \,
    \
    \label{rattention}
\end{align}
Then, in Equation~\ref{neighboraware}, a 1D convolution is applied to the sequence ${\mathrm{R\_attn}_{b,i}}$ to incorporate neighbor awareness $\Phi$. To handle boundary conditions, any ${\mathrm{R\_attn}_{b,i}}$ that refers to an index outside the valid range (i.e., $i < 1$ or $i > M$) is treated as zero.
\begin{align}
    \
    \mathrm{\Phi}_{b,i}
    = \sum_{d=-R}^{R} \tilde w_d \,\mathrm{R\_attn}_{b,i+d}
    \quad
    \
    \label{neighboraware}
\end{align}
The $\tilde{w}_d$ in Equation~\ref{neighboraware} denotes the weight at relative offset $d$ in a 1D convolution kernel of size $2R+1$, which is designed to incorporate information from neighboring tokens. As defined in Equation~\ref{kernal}, the kernel is constructed using a distance-based decay function, where the unnormalized weight at position $d$ is set to $1 / (|d| + 1)$. This ensures that closer neighbors contribute more than distant ones. The entire kernel is then normalized by dividing each weight by the sum over all offsets $u = -R, \dots, R$, so that the final weights $\tilde{w}_d$ sum to one. This formulation allows for a smooth and spatially-aware aggregation of token importance scores, where the influence of each neighboring token decays gradually with its distance from the center.
\begin{align}
    \
    \tilde w_d
    = \frac{\displaystyle{1}/({|d|+1})}
           {\displaystyle\sum_{u=-R}^{R}{1}/({|u|+1})}
    \,,
    \quad d=-R,\dots,R
    \
    \label{kernal}
\end{align}

The importance score $ \mathrm{\xi}_{b,i}$ of each token is computed using Equation~\ref{importance}, which combines neighbor awareness and the attention from the CLS token to each image token. The parameter $\alpha$ serves as a tunable weight to balance the contributions of the two components.
\begin{align}
    \
    \mathrm{\xi}_{b,i}
    = (1 - \alpha)\,\mathrm{cls\_attn}_{b,i}
    \;+\  \alpha\,\mathrm{\Phi}_{b,i}
    \,
    \
    \label{importance}
\end{align}

Tokens are sorted based on their importance scores. The engaged and attended tokens are kept for subsequent calculations, while the isolated and unattended tokens are aggregated into a single token and passed through the MLP for further calculation.

\subsection{Token Merging: MAT}
\vspace{5pt}
In Token Merging, we propose a merging strategy based on adjacent tokens, referred to as MAT (Merging Adjacent Tokens). The key idea behind this approach is that pixel values in natural images typically change smoothly, resulting in local semantic continuity. As a result, similar tokens are often spatially adjacent to each other. Based on this observation, instead of computing pairwise similarities across all tokens as done in ToMe’s BSM (Bipartite Soft Matching)~\cite{bolya2022token}, MAT focuses only on computing similarities between adjacent tokens. With the token sequence reordered using a Hilbert curve, adjacent tokens in the 1D sequence are more likely to remain adjacent in the 2D image space, making local similarity a meaningful basis for token merging.

The MAT module in Figure~\ref{fig:overview} illustrates the complete process of token merging. The similarity between each pair of tokens is computed as shown in Equation~\ref{similarity}, where $m_i\in R^C$ denotes the feature vector of the $i_{th}$ token. Various similarity metrics were evaluated, and the corresponding results are presented in Table~\ref{SimilarityAblation}. Let the total number of tokens be \textit{T}, and let \textit{p} denote the number of protected tokens (such as class tokens or distill tokens), which are excluded from the merging process. Accordingly, the number of adjacent token pairs considered for similarity computation is $N = T - p - 1$. In the similarity calculation, the vectors of tokens are individually normalized to unit vectors based on their $L2$ norms. Two similarity measures are considered: cosine similarity and symmetric KL divergence, with performance comparisons provided in Table~\ref{SimilarityAblation}. The resulting similarity score is denoted as $A_j$, representing the similarity between the $(p+j-1)_{th}$ and $(p+j)_{th}$ adjacent tokens. Thus, the set $\{A_1, A_2, \dots, A_N\}$ represents the similarity scores of all mergeable adjacent token pairs.
\begin{equation}
    A_j \;=\;\Biggl\langle \frac{m_{p+j-1}}{\|m_{p+j-1}\|_2},\;\frac{m_{p+j}}{\|m_{p+j}\|_2}\Biggr\rangle
    \;, \quad j = 1,\dots,N
    \label{similarity}
\end{equation}
A $topk$ selection is then applied to this set of scores to determine that the adjacent token pairs with the top $r$ similarities are to be merged. Specifically, as shown in Equation~\ref{sort}, the $Top$-$r$ similarities are selected, and their corresponding indexes $j_k$ are sorted in ascending order. 
\begin{equation}
    \{\,j_{(1)}, \dots, j_{(r)}\}
    = \mathrm{sort}\!\Bigl(\mathrm{topk}\bigl(\{A_j\}_{j=1}^N,\;r\bigr)\Bigr)
    \label{sort}
\end{equation}
The merge destination and source are subsequently determined based on the similarity index $j_k$. As defined in Equation~\ref{ds}, $D_k$ denotes the merge destination, which is the same as the similarity index $j_k$ (such as $Token_1$ in the MAT module of Figure~\ref{fig:overview}), and $S_k$ denotes the merge source (such as $Token_2$ and $Token_3$). Since only adjacent token pairs are considered for merging, $S_k$ is always equal to $D_{k+1}$.
\begin{equation}
    D_k = j_{(k)} = j_k, 
    \quad
    S_k = D_k + 1,
    \quad
    k = 1,\dots,r
    \label{ds}
\end{equation}
However, directly merging tokens pairwise according to $D_k$ and $S_k$ may lead to conflicts. For example, in MAT module of Figure~\ref{fig:overview}, if $D_1 = {Token_1}$, $S_1 = {Token_2}$, and $D_2 = {Token_2}$, $S_2 = {Token_3}$, then $Token_2$ is referenced by two different merging pairs simultaneously, which is not allowed. To address this issue, a mechanism of continuous interval detection and starting point forward filling is introduced.
\begin{align}
    F_k \;=\; D_{\displaystyle \max\bigl\{\,i \in \{1,\dots,k\}\mid i=1 \;\lor\; D_i - D_{i-1} > 1\bigr\}}
    \label{indicator}
\end{align}
As shown in Equation~\ref{indicator}, the new starting points of the merge intervals are identified by checking whether adjacent $D_k$ values are discontinuous (i.e., $D_k - D_{k-1} > 1$). For each $k$, the nearest $D_i$ that satisfies this discontinuity condition is selected and used as the final merge destination $F_k$. When $k = 1$, the set only contains $i = 1$, so $F_1 = D_1$. For example, in MAT module of Figure~\ref{fig:overview}, the original $D_k = [1, 2, 5]$, where $Token_1$ and $Token_2$ are continuous, resulting in the merge destination of $Token_2$ being updated to $Token_1$. Since the gap between $Token_5$ and the previous token is greater than 1, it is preserved as the starting point of a new interval. As a result, the final merge destinations are $F_k = [1, 1, 5]$, while the corresponding $S_k = [2, 3, 6]$ remain unchanged.
Merge source tokens $S_k$ sharing the same destination $F_k$ are aggregated by computing a weighted average together with $F_k$ itself, resulting in a single merged token. These merged tokens are then arranged into a new sequence for subsequent computations.

\subsection{Token Pruning and Merging: HyNAP}\label{TP&TM}

In this work, we integrate both token pruning and merging within the DiffRate~\cite{chen2023diffrate} framework. Specifically, we replace DiffRate’s original pruning strategy with our proposed Neighbor-Aware Pruning (NAP), which enhances importance estimation by incorporating local token interactions. For the merging stage, we retain DiffRate’s merging mechanism, where tokens are merged by importance, and pairwise similarities are computed using BSM within each group. This combined approach, termed Hybrid NAP (HyNAP), leverages the strengths of both pruning and merging while emphasizing neighbor-aware token selection.

Note that the MAT strategy is not employed in this setting, as the NAP already accounts for the influence of neighbor tokens, resulting in retained tokens that are spatially clustered rather than scattered. As a consequence, the tokens involved in subsequent merging tend to be locally distributed in the 2D image space. This naturally satisfies the original motivation for introducing MAT, reducing the search space and preserving spatial locality, thereby making the explicit application of MAT unnecessary in this case.

\begin{table}[]
\resizebox{\columnwidth}{!}{%
\begin{tabular}{clcc}
\hline
\begin{tabular}[c]{@{}c@{}}Similarity \\ Algorithm\end{tabular} &
  \multicolumn{1}{c}{Metrics} &
  Top-1 Acc (\%) &
  \begin{tabular}[c]{@{}c@{}}Throughput\\ (imgs/s)\end{tabular} \\ \hline
\multirow{3}{*}{\begin{tabular}[c]{@{}c@{}}Cosine \\ Similarity\end{tabular}} &
  \textbf{K (mean)} &
  \textbf{78.4} &
  \textbf{1156} \\
 & X           & 78.2 & 1153 \\
 & K (L2 norm) & 78.2 & 1153 \\ \hline
\multirow{3}{*}{\begin{tabular}[c]{@{}c@{}}KL \\ Divergence\end{tabular}} &
  K (mean) &
  78.3 &
  1130 \\
 & X           & 78.3 & 1124 \\
 & K (L2 norm) & 78.2 & 1128 \\ \hline
\end{tabular}%
}
\caption{Similarity metrics and algorithm. $K (mean)$ represents the average of the key vectors of each head; $X$ represents the L2 norm of each token feature value; $K (norm)$ represents the L2 norm of the key vector. The evaluation is performed on DeiT-S/224 with halved flops.}
\label{SimilarityAblation}
\end{table}

\section{Experiments}\label{sec:experiments}
\subsection{Implementation Details}

During the evaluation process, all models are used in an off-the-shelf manner. NAP and MAT can be directly applied to the corresponding ViT models without additional training. In this work, the original ViT and its variants (such as DeiT~\cite{touvron2021training} and ViT (MAE)~\cite{he2022masked}) are used as baseline models, and several classic token merging and pruning methods are selected for comparison. For the convenience of comparison, the parameter settings of MAT and NAP, including the merging rate, pruning rate, and the layers for token reduction, follow the same practices as ToMe, EViT, and DiffRate. All experiments are conducted on an RTX 3080 GPU. Due to hardware memory limitations, larger models require a reduced batch size. To ensure fair evaluation, the batch size is uniformly set to 16 during evaluation. The dataset used in the experiments is ImageNet-1K~\cite{deng2009imagenet}, and Top-1 Accuracy is adopted as the accuracy metric, while FLOPs and throughput are used to measure computational efficiency. All code and data will be open-sourced upon acceptance.

\subsection{Results Analysis}
\subsubsection{Token pruning}
\begin{figure*}[t]
    \centering
    \subcaptionbox{\textbf{Token pruning comparison.} The baseline is DeiT-B/384, and the result is without any finetune. With the same efficiency improvement, NAP has higher accuracy.\label{fig:hilvsevit}}[0.3\linewidth]{%
        \includegraphics[width=\linewidth]{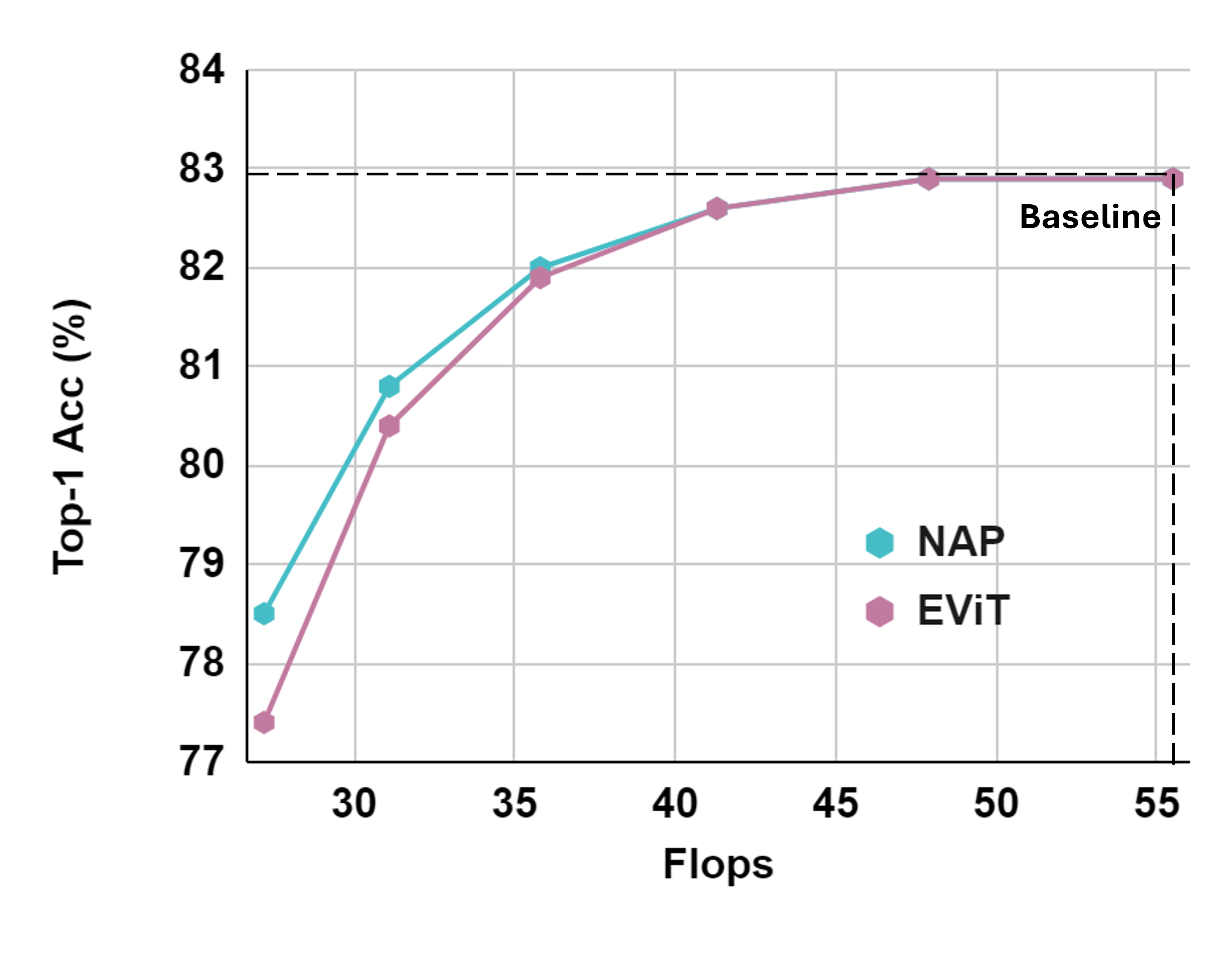}
    }
    \hspace{0.02\linewidth}
    \subcaptionbox{\textbf{Token merging comparison.} The baseline is ViT-L/224 (MAE), and the results are without any finetune. When Flops is reduced more, ToMe has higher accuracy, and vice versa, MAT has higher accuracy.\label{fig:hilvstome}}[0.3\linewidth]{%
        \includegraphics[width=\linewidth]{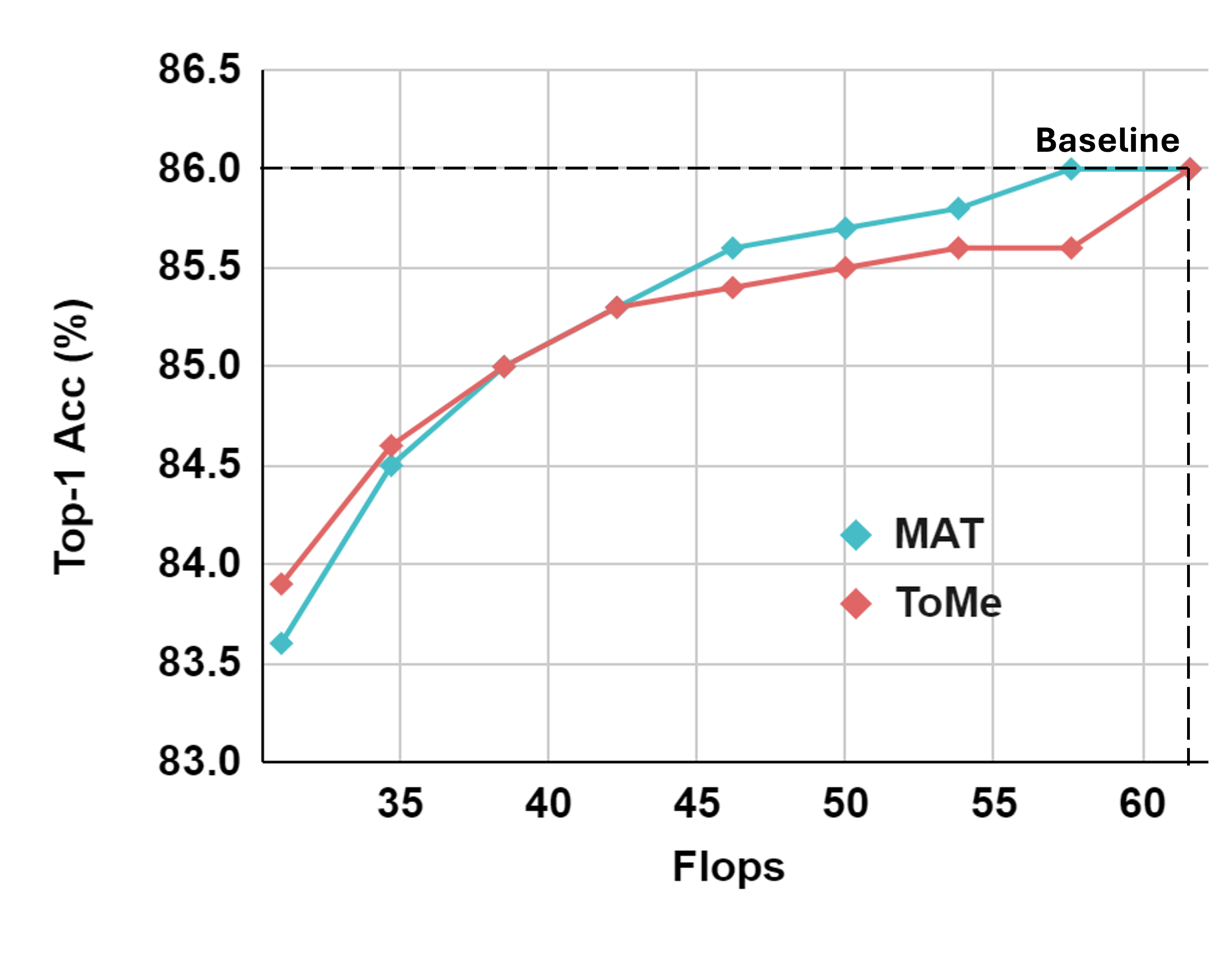}
    }
    \hspace{0.02\linewidth}
    \subcaptionbox{\textbf{The comparison between NAP and MAT.} The baseline is DeiT-B/224, and the results are without any finetune. The size of the point represents the throughput. The larger the size of the point, the higher the throughput.\label{fig:hilmvsp}}[0.3\linewidth]{%
        \includegraphics[width=\linewidth]{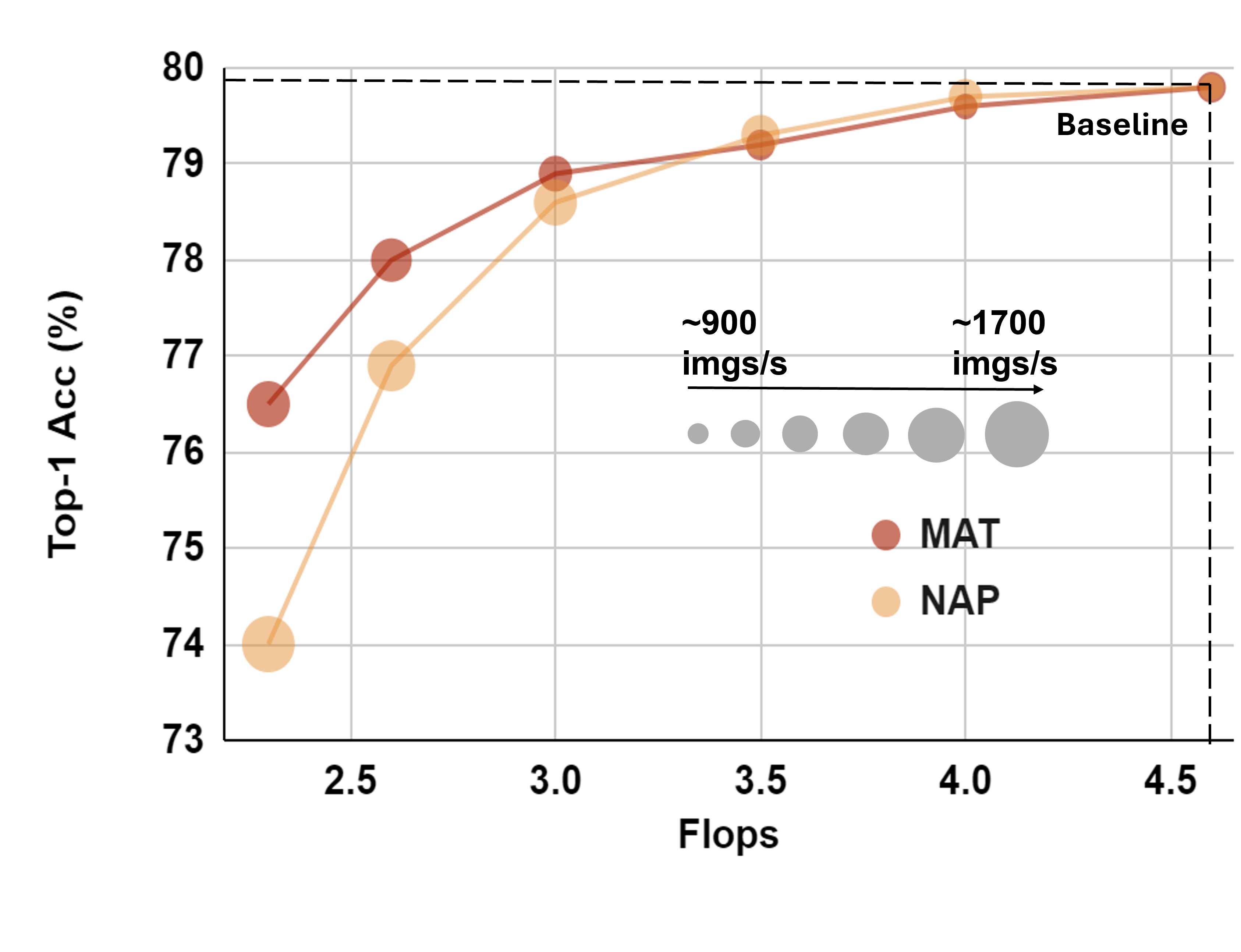}
    }
    \caption{Comparison of NAP, MAT, and other token reduction methods at different token reduction rates} 
    \label{hilvs}
\end{figure*}
In Table~\ref{pruningresults}, we compare the performance of different token pruning methods on various DeiT models. Here, NAP refers to directly applying pruning to DeiT-S and DeiT-B without any training. In Table~\ref{ftpruningresults}, DynamicViT* (DyViT*)~\cite{rao2021dynamicvit}, Evo-ViT~\citep{xu2022evo}, and EViT*~\cite{chen2023sparsevit} involve applying their respective pruning techniques to the baseline models followed by fine-tuning, while NAP* applies the proposed pruning method directly on EViT*. Therefore, NAP does not introduce any additional parameters or FLOPs to the model. It merely changes the way important tokens are selected, making it a plug-and-play solution. As shown, NAP on DeiT-S/224 reduces FLOPs by $50\%$ and improves throughput by $77\%$, with only a $1.2\%$ drop in accuracy. Due to the introduction of neighbor-awareness, the throughput is slightly lower compared to EViT, but this difference becomes less noticeable in larger models. At higher resolutions, such as on DeiT-B/384 in Table~\ref{pruningresults}, both NAP and EViT nearly double in efficiency, with NAP reducing accuracy by $4.4\%$ while EViT causes a larger drop of $5.5\%$.

\begin{table}[htp]
\resizebox{\columnwidth}{!}{%
\begin{tabular}{@{}clccc@{}}
\toprule
Model                       & Method      & Flops          & Top-1 Acc (\%)      & Throughput \\ \hline
\multirow{5}{*}{\begin{tabular}[c]{@{}c@{}}DeiT-S\\ /224\end{tabular}} & Baseline    & 4.6G           & 79.8          & 915 imgs/s        \\ \cmidrule(l){2-5} 
                            & DyViT~\cite{rao2021dynamicvit}  & 2.2G           & 51.7          & 1585 imgs/s      \\ 
                            & EViT~\cite{chen2023sparsevit}        & 2.3G           & 73.8          & 1660 imgs/s       \\
                            & Evo-ViT~\cite{xu2022evo}        & 2.3G           & 72.1          & 1403 imgs/s       \\
                            & NAP (ours)  & 2.3G  & 74.0 & 1600 imgs/s        \\ \midrule
\multirow{5}{*}{\begin{tabular}[c]{@{}c@{}}DeiT-B\\ /224\end{tabular}} & Baseline    & 17.6G          & 81.8          & 283 imgs/s        \\ \cmidrule(l){2-5}
                            & DyViT~\cite{rao2021dynamicvit}  & 11.6G           & 75.5          & 420 imgs/s      \\ 
                            & EViT~\cite{chen2023sparsevit}        & 11.5G          & 80.5          & 440 imgs/s        \\
                            & Evo-ViT~\cite{xu2022evo}        & 11.6G           & 79.1          & 415 imgs/s       \\
                            & NAP (ours)  & 11.5G & 80.5 & 434 imgs/s         \\ \midrule
\multirow{3}{*}{\begin{tabular}[c]{@{}c@{}}DeiT-B\\ /384\end{tabular}} & Baseline    & 55.5G           & 82.9          & 89 imgs/s         \\ \cmidrule(l){2-5} 
                            & EViT~\cite{chen2023sparsevit}        & 27.2G           & 77.4          & 175 imgs/s        \\
                            & NAP (ours)  & 27.2G  & 78.5 & 173 imgs/s        \\ \bottomrule
\end{tabular}%
}
\caption{\textbf{Token pruning results.}The results are obtained by directly applying different token pruning methods on the baseline without any fine-tuning. Because R and $\alpha$ in Equation~\ref{kernal} and~\ref{importance} are adjustable. In DeiT-S/224 and DeiT-B/224, they are set to $R=3$, $\alpha=0.1$. In DeiT-B/384, they are set to R=3, $\alpha= 0.95$.}
\label{pruningresults}
\end{table}

\begin{table}[t]
\resizebox{\columnwidth}{!}{%
\begin{tabular}{@{}clccc@{}}
\toprule
Model                       & Method      & Flops          & Top-1 Acc (\%)      & Throughput \\ \hline
\multirow{5}{*}{\begin{tabular}[c]{@{}c@{}}DeiT-S\\ /224\end{tabular}} & Baseline    & 4.6G           & 79.8          & 915 imgs/s        \\ \cmidrule(l){2-5} 
                            & DyViT*~\cite{rao2021dynamicvit} & 2.2G           & 77.5          & 1585 imgs/s       \\
                            & Evo-ViT*~\cite{xu2022evo}        & 2.3G           & 77.7          & 1403 imgs/s       \\
                            & EViT*~\cite{chen2023sparsevit}       & 2.3G           & 78.5          & 1670 imgs/s       \\
                            & NAP* (ours) & 2.3G  & 78.6 & 1622 imgs/s        \\  \hline
\end{tabular}%
}
\caption{\textbf{Token pruning fine-tuned results.} Both DyViT* and EViT* apply the corresponding pruning technique to the baseline and then perform fine-tuning. The plug-and-play NAP* directly applies EViT* without any fine-tuning.}
\label{ftpruningresults}
\end{table}

To compare NAP and EViT at different keeping ratios, we present their evaluation results on DeiT-S/224 under varying keeping rates in Figure~\ref{fig:hilvsevit}. The results demonstrate that NAP outperforms EViT at lower keeping ratios, confirming the benefits of neighbor-awareness in token pruning.

\subsubsection{Token merging}
In Table~\ref{tokenmergingresults}, we compare the performance of MAT and other token merging techniques on various DeiT models. Under the current merging ratio settings, MAT performs better than LoTM and K-Medoids, and exhibits comparable performance to ToMe across these different models. Fine-tuning has the potential to enhance accuracy even more. MAT does not introduce additional computational overhead or parameters, as it only modifies the similarity computation method and merging strategy. However, due to connected-region detection and forward filling, as described in Equation~\ref{ds} and Equation~\ref{indicator}, MAT results in slightly lower throughput compared to ToMe. Nevertheless, this difference diminishes as the model size increases.

\begin{table}[htp]
\resizebox{\columnwidth}{!}{%
\begin{tabular}{@{}clccc@{}}
\toprule
Model & Method         & Flops  & Top-1 Acc (\%) & Throughput \\ \midrule
\multirow{5}{*}{\begin{tabular}[c]{@{}c@{}}DeiT-S\\ /224\end{tabular}}        & Baseline & 4.6G    & 79.8 & 915 imgs/s \\ \cmidrule(l){2-5} 
      & ToMe~\cite{bolya2022token}           & 3.3G   & 79.3     & 1044 imgs/s       \\
      & K-Medoids~\cite{marin2023token}           & 3.3G   & 78.7     & --       \\
      & LoTM~\cite{haroun2024leveraging}           & 3.3G   & 79.2     & --       \\
      & MAT (ours) & 3.3G   & 79.3     & 980 imgs/s       \\ \midrule
\multirow{5}{*}{\begin{tabular}[c]{@{}c@{}}DeiT-B\\ /224\end{tabular}}        & Baseline & 17.6G  & 81.8 & 283 imgs/s \\ \cmidrule(l){2-5} 
      & ToMe~\cite{bolya2022token}           & 12.6G  & 80.9     & 367 imgs/s         \\
      & K-Medoids~\cite{marin2023token}           & 12.3G   & 80.0     & --       \\
      & LoTM~\cite{haroun2024leveraging}           & 12.3G   & 80.0     & --       \\
      & MAT (ours) & 12.6G  & 80.9     & 354 imgs/s        \\ \midrule
\multirow{3}{*}{\begin{tabular}[c]{@{}c@{}}DeiT-B\\ /384\end{tabular}}        & Baseline & 55.5G  & 82.9 & 85 imgs/s  \\ \cmidrule(l){2-5} 
      & ToMe~\cite{bolya2022token}           & 37.0G  & 82.4     & 120 imgs/s        \\
      & MAT (ours) & 37.0G  & 82.5     & 120 imgs/s          \\ \bottomrule
\end{tabular}%
}
\caption{\textbf{Token merging results.} All results are re-implemented and without fine-tuning.}
\label{tokenmergingresults}
\end{table}

To compare MAT and ToMe at different merging ratios, Figure~\ref{fig:hilvstome} presents their performance on ViT-L/224 (MAE) under varying merging rates. The results show that at low FLOPs (i.e., high merging ratios), ToMe achieves better performance, whereas MAT outperforms at lower merging ratios. This is primarily because ToMe benefits from its global search capability, while MAT emphasizes locality. When the number of tokens to be merged is small, prioritizing the merging of locally adjacent tokens often leads to better preservation of spatial coherence. However, at high merging ratios, global search can better find and merge semantically similar tokens that local methods might miss. This observation suggests a limitation of MAT: since it only computes similarity between adjacent tokens, it may miss semantically similar tokens that are not spatially close. Additionally, while Hilbert-curve ordering ensures consecutive sequence tokens are spatially adjacent, spatial adjacency does not guarantee sequence consecutiveness. A potential improvement could be expanding the search range to small blocks, where similarity is computed among tokens within each block. This approach would maintain spatial locality while also capturing semantically similar tokens that are not immediate neighbors, thereby improving merging effectiveness at higher merging ratios.
\subsubsection{Token pruning and merging}
In Figure~\ref{fig:hilmvsp}, we present a comparative analysis of MAT and NAP on DeiT-S/224. The results demonstrate that MAT achieves superior performance when aggressive token reduction is required, while NAP exhibits better performance with more conservative token reduction. This occurs because MAT merges similar tokens to preserve local context, whereas NAP discards tokens completely, resulting in greater information loss at high reduction rates. In terms of throughput, NAP has always been in the lead. This observation provides clear guidance for selecting the appropriate method based on the desired token reduction ratio.

\begin{table}[htp]
\resizebox{\columnwidth}{!}{%
\begin{tabular}{@{}clccc@{}}
\toprule
Model                   & Method          & Flops        & Top-1 Acc (\%) & Throughput    \\ \midrule
\multirow{10}{*}{\begin{tabular}[c]{@{}c@{}}DeiT-S\\ /224\end{tabular}}  & Baseline        & 4.6G          & 79.8           & 915 imgs/s           \\ \cmidrule(l){2-5} 
                        & ToMe~\cite{bolya2022token}            & 2.9G            & 78.7           & 1194 imgs/s          \\
                        & LoTM~\cite{haroun2024leveraging}            & 2.9G            & 78.8           & --           \\
                        & DyViT~\cite{rao2021dynamicvit}            & 3.0G            & 67.4           & 1255 imgs/s          \\
                        & Evo-ViT~\cite{xu2022evo}            & 3.0G            & 77.4           & 1213 imgs/s          \\
                        & ATS~\cite{fayyaz2022adaptive}            & 3.0G            & 79.2           & 736 imgs/s          \\
                        & GTP-ViT~\cite{xu2024gtp}            & 3.0G            & 79.1           & 1281 imgs/s          \\
                        & EViT~\cite{chen2023sparsevit}            & 3.0G            & 78.5           & 1324 imgs/s \\
                        & DiffRate~\cite{chen2023diffrate}        & 2.9G            & 79.6           & 1174 imgs/s          \\
                        & Zero-TP~\cite{wang2024zero}        & 3.0G            & 79.4           & 1131 imgs/s          \\
                        & HyNAP (ours) & 2.9G & 79.6  & 1100 imgs/s          \\ \midrule
\multirow{10}{*}{\begin{tabular}[c]{@{}c@{}}DeiT-B\\ /224\end{tabular}} & Baseline        & 17.6G           & 81.8           & 283 imgs/s           \\ \cmidrule(l){2-5} 
                        & ToMe~\cite{bolya2022token}            & 8.8G            & 77.5           & 528 imgs/s           \\
                        & DyViT~\cite{rao2021dynamicvit}            & 8.8G            & 50.2           & 560 imgs/s          \\
                        & Evo-ViT~\cite{xu2022evo}            & 8.8G           & 60.6           & 496 imgs/s          \\
                        & ATS~\cite{fayyaz2022adaptive}            & 8.8G            & 79.6           & 336 imgs/s          \\
                        & GTP-ViT~\cite{xu2024gtp}            & 8.8G            & 78.3           & 565 imgs/s          \\
                        & EViT~\cite{chen2023sparsevit}            & 8.7G            & 75.1           & 570 imgs/s  \\
                        & DiffRate~\cite{chen2023diffrate}        & 8.7G            & 79.0           & 509 imgs/s           \\
                        & HyNAP (ours) & 8.7G & 79.3  & 504 imgs/s           \\ \midrule
\multirow{5}{*}{\begin{tabular}[c]{@{}c@{}}ViT-B\\ /224\\ (MAE)\end{tabular}} & Baseline        & 17.6G           & 83.6           & 280 imgs/s           \\ \cmidrule(l){2-5}                                               & ToMe~\cite{bolya2022token}            & 8.8G            & 78.9           & 535 imgs/s           \\
                        & EViT~\cite{chen2023sparsevit}            & 8.7G            & 75.1           & 572 imgs/s  \\
                        & DiffRate~\cite{chen2023diffrate}        & 8.7G            & 79.9           & 508 imgs/s           \\
                        & HyNAP (ours) & 8.7G & 80.1  & 503 imgs/s           \\ \midrule
\multirow{5}{*}{\begin{tabular}[c]{@{}c@{}}ViT-L\\ /224\\ (MAE)\end{tabular}} & Baseline        & 61.6G           & 85.7           & 89 imgs/s           \\ \cmidrule(l){2-5}                                               & ToMe~\cite{bolya2022token}            & 42.3G            & 85.4           & 125 imgs/s           \\
                        & EViT~\cite{chen2023sparsevit}            & 39.6G            & 85.1           & 147 imgs/s  \\
                        & DiffRate~\cite{chen2023diffrate}        & 42.3G            & 85.6           & 120 imgs/s           \\
                        & HyNAP (ours) & 42.3G & 85.7  & 120 imgs/s           \\ \midrule                      
\end{tabular}%
}
\caption{Comparison of HyNAP and other token reduction methods in terms of accuracy and efficiency.}
\label{Allresults}
\end{table}

For the combined pruning-and-merging approach, we adopt the token reduction framework from DiffRate, specifically integrating NAP with the merging component of DiffRate. Table~\ref{Allresults} presents a comprehensive comparison of the performance and efficiency between the HyNAP approach and other token reduction methods. On DeiT-S/224, HyNAP achieves a $37\%$ reduction in FLOPs and a $20\%$ improvement in throughput while maintaining competitive accuracy with only a $0.2\%$ accuracy drop. Although the throughput improvement is slightly lower than that of some alternatives, HyNAP consistently outperforms most SOTA token reduction methods at comparable computational budgets.
\begin{figure}[t]
  \centering
  \begin{subfigure}[b]{0.45\textwidth}
    \centering
    \includegraphics[width=\linewidth]{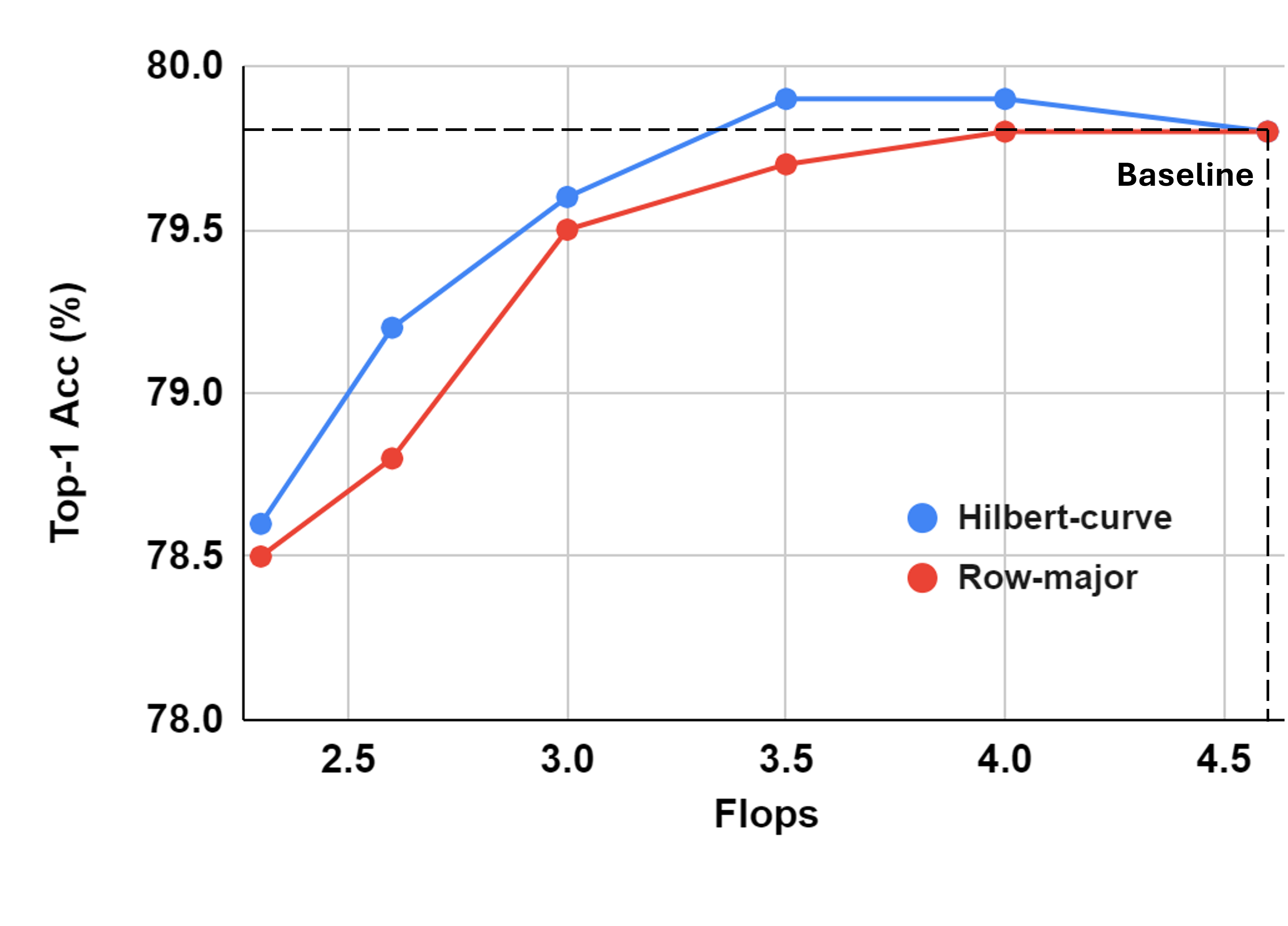}
    \caption{\textbf{NAP performance under different ordering schemes.} Using EViT* as the baseline, the performance in NAP when applying Hilbert-curve reordering versus conventional row-major ordering is analyzed.}
    \label{hilbertvsrow_prune}
  \end{subfigure}\hfill
  \begin{subfigure}[b]{0.45\textwidth}
    \centering
    \includegraphics[width=\linewidth]{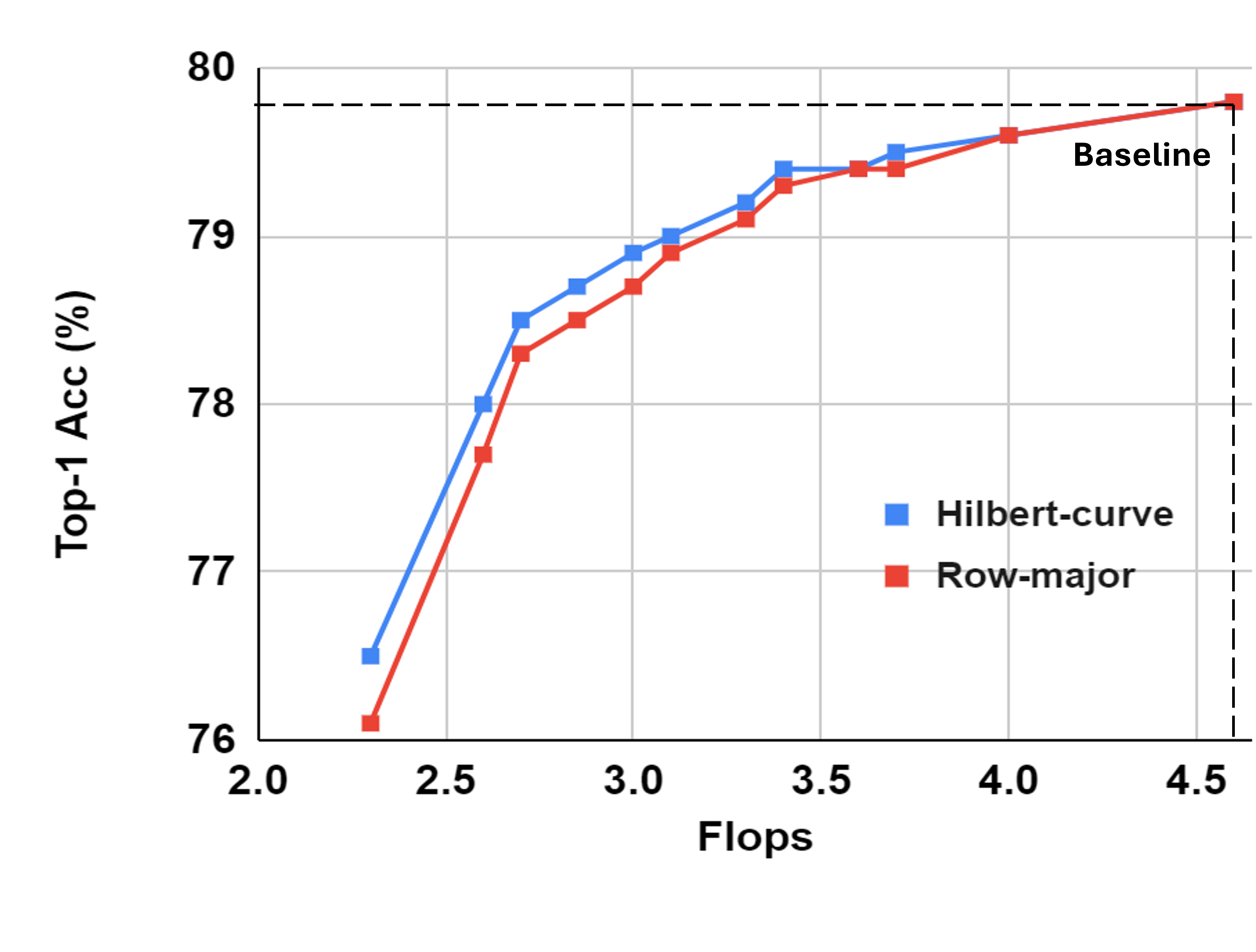}
    \caption{\textbf{MAT performance under different ordering schemes.} DeiT-B/224 is the baseline model.}
    \label{hilbertvsrow_merge}
  \end{subfigure}
  \caption{Performance under different ordering schemes.}
  \label{hilbertvsrow}
\end{figure}
Notably, on DeiT-B/224, HyNAP delivers more significant improvements: a $50\%$ reduction in FLOPs and a $78\%$ increase in throughput, with merely a $2.5\%$ accuracy degradation. For ViT-L/224 (MAE), HyNAP reduces FLOPs by $31\%$ without sacrificing any accuracy, and achieves a $35\%$ speedup. Compared to alternative approaches under these settings, HyNAP achieves comparable efficiency gains while maintaining better model accuracy. These results suggest that our method offers a favorable trade-off between efficiency and accuracy for certain model configurations.
\subsubsection{The comparison of different token ordering schemes}
Figure~\ref{hilbertvsrow} compares the performance of Hilbert curve reordering and row-major order in NAP and MAT. For token pruning evaluation, it is conducted on the EViT*, where different token orderings alter neighbor tokens and thus influence neighbor-awareness. As shown in Figure~\ref{hilbertvsrow_prune}, computing neighbor-awareness based on Hilbert curve reordering allows NAP to achieve an accuracy increase of $0.1\%$ to $0.4\%$. Similarly, Hilbert curve reordering alters adjacent token pairs, thereby changing the search space for similar tokens. The results in Figure~\ref{hilbertvsrow_merge} demonstrate that Hilbert curve reordering also improves the performance of MAT by up to $0.4\%$, outperforming the original row-major order at all merging ratios. Notably, Hilbert curve reordering incurs negligible computational cost since each feature map size only needs to be reordered once, with results cached for reuse.

\section{Conclusion}\label{sec:conclusion}
In this work, we propose neighbor-aware token reduction approaches: NAP and MAT, both inspired by the Hilbert Curve, to enhance the efficiency of ViTs. By introducing neighbor-awareness after Hilbert curve reordering, we adaptively apply token reduction strategies: MAT merges adjacent and semantically similar tokens, while NAP selectively preserves tokens based on both neighbor-awareness and attention scores, retaining highly attended tokens and discarding isolated or less relevant ones.
Extensive experiments on various ViT architectures show that NAP and MAT achieve a competitive balance between performance and efficiency compared to existing token reduction methods. 
However, MAT exhibits limitations under high merging ratios, as it may miss more suitable token pairs due to restricted local similarity matching. This issue could be mitigated by broadening the local search scope for similar tokens. NAP and MAT offer a practical and effective framework for efficient Transformer design, highlighting the importance of spatial continuity and local context in future research.

\section{Acknowledgment}
This work was supported in part by the NSF under award \#2245729.

\bigskip
\bibliography{aaai2026}

\end{document}